\documentclass[10pt, a4paper]{article}

\usepackage[final]{lrec-coling2024} 

\title{\textbf{Can Large Language Models Discern Evidence for Scientific Hypotheses? Case Studies in the Social Sciences}
}
\usepackage{multirow}
\usepackage{multicol}
\usepackage{tablefootnote}
\usepackage{caption}
\usepackage{subcaption}
\usepackage{booktabs}

\name{Sai Koneru$^1$, Jian Wu$^2$, Sarah Rajtmajer$^1$} 

\address{$^1$ Pennsylvania State University, State College, PA \\
$^2$ Old Dominion University, Norfolk, VA \\
     \{sdk96, smr48\}@psu.edu, j1wu@odu.edu\\}

\abstract{
Hypothesis formulation and testing are central to empirical research. A strong hypothesis is a best guess based on existing evidence and informed by a comprehensive view of relevant literature. However, with exponential increase in the number of scientific articles published annually, manual aggregation and synthesis of evidence related to a given hypothesis is a challenge. Our work explores the ability of current large language models (LLMs) to discern evidence in support or refute of specific hypotheses based on the text of scientific abstracts. We share a novel dataset for the task of \emph{scientific hypothesis evidencing} using community-driven annotations of studies in the social sciences. We compare the performance of LLMs to several state of the art methods and highlight opportunities for future research in this area. Our dataset is shared with the research community: \url{https://github.com/Sai90000/ScientificHypothesisEvidencing.git}
 \\ \newline \Keywords{Large Language Models, Natural Language Understanding, Scientific Hypothesis Evidencing}}

\begin{document}

\maketitleabstract

\section{Introduction}
 
Translating scholarly research findings into actionable, evidence-based impacts relies on iterative refinement for robust understanding of a given phenomenon across multiple studies, contexts, etc. The sequential approach to scientific interrogation is also at the heart of null hypothesis significance testing. Namely, a hypothesis is an informed theory, or an \emph{educated guess}, based on available information and prior findings \citep{wald1992sequential}. As such, synthesis and understanding of current literature is essential to study planning and to efficient research more broadly. Yet, scholarly databases fail to aggregate, compare, contrast, and contextualize existing studies in a way that allows comprehensive review of the relevant literature in service to a targeted research question. In part, this is because the sheer volume of published work is difficult to navigate and the narrative format through which most empirical work is reported was not envisioned with machine readability in mind. 

Work in the areas of natural language processing (NLP) and natural language understanding (NLU) has emerged to address various challenges related to synthesizing scientific findings. 
Automated approaches for \textit{fact-checking}~\citep{guo2022survey}, for example, have received significant attention in the context of misinformation and disinformation. This task aims to assess the accuracy of a factual claim based on a literature~\citep{vladika2023scientific}.  
What remains a gap, however, are methods to determine whether a research question is addressed within a paper based on its abstract, and if so, whether the corresponding hypothesis is supported or refuted by the work. In this work, we propose this task as \emph{scientific hypothesis evidencing} (SHE). 

\begin{table}[]
\small
  \centering
  \begin{tabular}{|p{0.95\columnwidth}|}
     \toprule
     \textbf{Research question (from the review).} Is there an association between social media use and bad mental health outcomes?\\
    \midrule
     \textbf{Abstract.} Although studies have shown that increases in the frequency of social media use may be associated with increases in depressive symptoms of individuals with depression, the current study aimed to identify specific social media behaviors related to major depressive disorder (MDD). Millennials (N = 504) who actively use Facebook, Twitter, Instagram, and/or Snapchat participated in an online survey assessing major depression and specific social media behaviors. Univariate and multivariate analyses were conducted to identify specific social media behaviors associated with the presence of MDD. The results identified five key social media factors associated with MDD. Individuals who were more likely to compare themselves to others better off than they were (p = 0.005), those who indicated that they would be more bothered by being tagged in unflattering pictures (p = 0.011), and those less likely to post pictures of themselves along with other people (p = 0.015) were more likely to meet the criteria for MDD. Participants following 300 + Twitter accounts were less likely to have MDD (p = 0.041), and those with higher scores on the Social Media Addiction scale were significantly more likely to meet the criteria for MDD (p = 0.031). Participating in negative social media behaviors is associated with a higher likelihood of having MDD. Research and clinical implications are considered. \\
     \midrule
     \textbf{Hypothesis (declarative).} There is an association between social media use and bad mental health outcomes.\\
     \midrule
     \textbf{Label.} Entail\\
    \bottomrule
  \end{tabular}
  \caption{An example from the training dataset containing a paper's \emph{abstract}, a \emph{hypothesis} of interest, and corresponding \textit{label} identifying the relationships between the hypothesis and the abstract.}
  \label{tab:example}
\end{table}

Notably, grassroots efforts to usefully assemble the literature have popped up, e.g., in the form of shared Google docs, contributed to by authors of related work and socialized primarily via Twitter \citep{haidt2022social}. In these documents, authors synthesize existing work that tests closely-related hypotheses or a similar research question, e.g., \emph{Does social media cause political polarization?} Chapters within these collaborative documents add additional structure, highlighting studies with similar outcomes or similar experimental settings. 

In this work, we study whether and to what extent state-of-the-art NLU and large language models can supplant manual expert-driven collaborative meta-analyses, or parts thereof, in service to discerning hypotheses and primary findings from scientific abstracts. We focus on the social sciences due to the availability of high quality datasets annotated by domain experts. Our work makes the following primary contributions: 

\begin{enumerate}
  \item We propose the SHE task--the identification of evidence from the abstract of a scientific publication in support or refute of a hypothesis;
  \item We build and share a benchmark dataset for SHE using expert-annotated collaborative literature reviews; 
   {\item Using this dataset, we evaluate performance of state of the art transfer learning and large language models (LLMs) for the SHE task.}
\end{enumerate}

\noindent Our findings suggest that this task is challenging for current NLU and that LLMs do not seem to perform better than traditional language models and transfer learning models. We offer perspectives and suggestions for the path forward.

\section{Related Work}

The task of \textit{scientific claim verification} is treated either: (1) as a natural language inference (NLI) problem using deep neural networks trained on human-annotated datasets~\citep{khot2018scitail,wadden2022scifact}; or, (2) as a classification problem using a joint claim-evidence representation \citep{oshikawa2018survey}.
In support of these efforts, multiple \textit{claim verification} datasets have been proposed as benchmarks for the community, e.g., for topics in biomedical sciences \citep{wadden2020fact, wadden2022scifact}, public health \citep{kotonya2020explainable, sarrouti2021evidence,saakyan2021covid} and environment \citep{diggelmann2020climate}. Examples of the NLI datasets include the Stanford Natural Language Inference (SNLI) dataset~\citep{bowman2015snli} and the Allen AI's SciTail dataset~\citep{khot2018scitail}. SNLI contains about 550,000 premise-hypothesis pairs. The premises were derived from image captions and hypotheses were created by crowdworkers. SNLI was the first NLI corpus to see encouraging results from neural networks. The SciTail dataset contains 27,000 premise-hypothesis pairs created from multiple-choice science exams and web sentences. Examples of the claim-evidence representations include the SciFact dataset \citep{wadden2020fact} and its extension -- SciFact-Open \citep{wadden2022scifact}. SciFact contained about 1.4K scientific claims and a search corpus of about 5K abstracts that provided either supporting or refuting evidence for each claim. The claims in SciFact-open were extracted from the citation context of papers in biomedical sciences including 279 claims verified against a search corpus of 500K abstracts. 

LLMs are trained on large datasets sourced from the internet representing a wide spectrum of both general and domain knowledge. They have shown remarkable performance across a range of NLU tasks such as reading comprehension, and question answering~\citep{liang2022holistic}. The SHE task offers a distinctive opportunity to assess these models in the context of scientific research domain expertise, thereby enabling reasoning abilities compared to those of human experts. 

The SHE problem formulation is distinct from the hypothesis and premise pairs encountered in conventional NLI tasks \citep{bowman2015snli}. The language used in scientific publications contains domain-specific terminology which is different from the the premise-hypothesis pairs in general scientific domains (e.g., SNLI \citep{bowman2015snli}). Furthermore, abstracts from scientific articles contain numerical data that is often not present in traditional NLP datasets.
\begin{figure}[t]
 \centering
 \includegraphics[width=\linewidth]{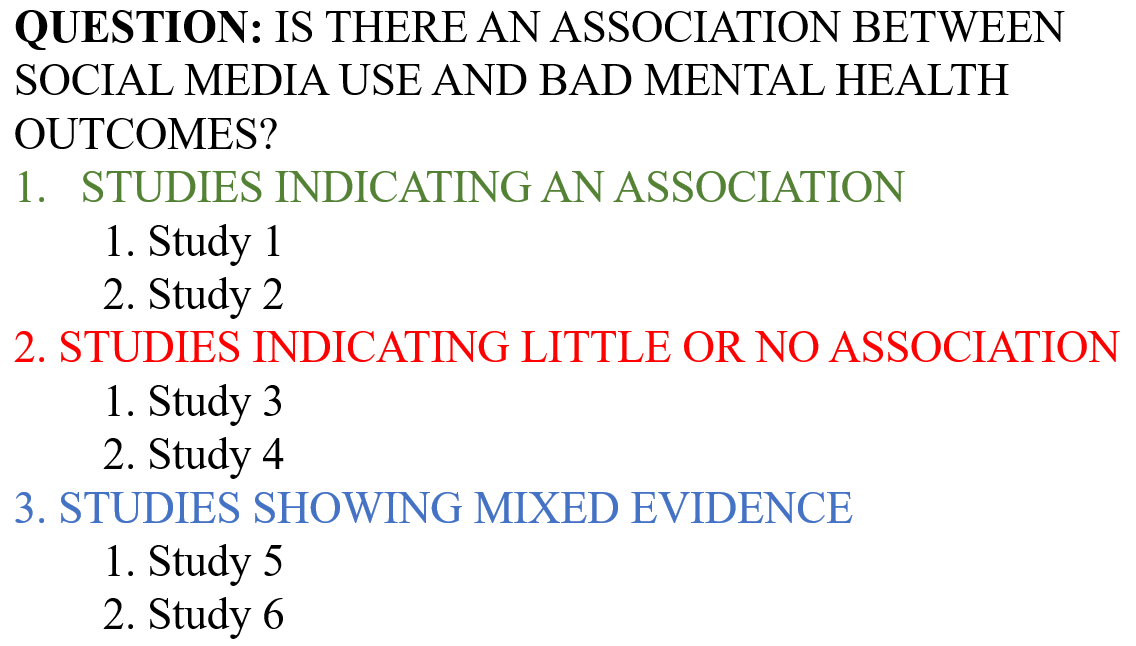}
 \caption{Exemplar collaborative review document structure for one question.}
 \label{fig:docstructure}
 \vspace{-0.2cm}
\end{figure}
\section{Problem Definition}
Scientific hypothesis evidencing (SHE) is defined as the identification of the association between a given declarative hypothesis and a relevant abstract. This association can be labeled either \textit{entailment}, \textit{contradiction}, or \textit{inconclusive}. The complexity of the task arises from contextual reasoning. For example, in Table \ref{tab:example}, identifying the relationship between the abstract and hypothesis provided requires the model to reason that \textit{depressive disorder} is a \textit{bad mental health outcome}, \textit{usage of Twitter, Facebook, Instagram, Snapchat} is \textit{use of social media} leading to \textit{higher likelihood of major depressive disorder} and hence the relationship is \textit{entailment}. In SHE, the hypotheses or research questions are typically expressed at a higher level of abstraction than the evidence provided within the abstract. 
In this work, we assume that the hypothesis in each hypothesis-abstract pair is addressed by the paper in question and focus on identification of relations between hypotheses and abstracts. Identifying evidence about an arbitrary hypothesis from a literature database is a bigger challenge, usually involving an information retrieval component. This would require a larger corpus of labeled documents as ground truth~\citep{wadden2022scifact, pradeep2021scientific}. 

\begin{table*}[t]
\centering
\small
\selectfont{
\begin{tabular}{l|llllll|lll}
\toprule
  \textbf{Topic} & \textbf{Hyp.} & \textbf{Art.} & \textbf{Tri.} & \textbf{Ent.} & \textbf{Cont.} & \textbf{Inc.} & \textbf{Train} & \textbf{Dev.} & \textbf{Test}\\
  \midrule
{Adolescent mood disorders} & 4 & 34 & 37 & 36 & 0 & 1 & 27 & 10 & 0\\
{Adolescent mental illness crisis } & 8 & 40 & 40 & 35 & 0 & 5 & 25 & 8 & 7\\
{Changes in cognitive ability} & 1 & 12 & 13 & 11 & 0 & 2 & 10 & 3 & 0\\
{Digital gambling and mental health} & 1 & 3 & 3 & 0 & 3 & 0 & 2 & 1 & 0\\
{Free play and mental health} & 5 & 36 & 37 & 23 & 13 & 1 & 17 & 8 & 12\\
{Online communities and adolescent health} & 2 & 3 & 3 & 1 & 0 & 2 & 1 & 1 & 1\\
{Phone free schools} & 5 & 37 & 38 & 8 & 26 & 4 & 21 & 8 & 9\\
{Porn use and adolescent health} & 6 & 47 & 47 & 14 & 24 & 9 & 30 & 10 & 7\\
{Social media and mental health} & 14 & 222 & 232 & 178 & 48 & 6 & 142 & 38 & 52 \\
{Social media and political dysfunction} & 9 & 144 & 152 & 67 & 40 & 45 & 82 & 35 &35\\
{Video game use and adolescent health} & 9 & 30 & 32 & 18 & 9 & 5 & 23 & 4 & 5\\
{Gen Z Phone-Based Childhood} & 2 & 3 & 3 & 2 & 0 & 1 & 2 & 1 & 0\\
\midrule
\textbf{Total} & 69 & 602 & 637 & 393 & 163 & 81\\
\textbf{Train} & 59 & 370 & 382 & 243 & 92 & 47\\
\textbf{Dev.} & 46 & 127 & 127 & 79 & 28 & 20\\
\textbf{Test} & 35 & 126 & 128 & 71 & 43 & 14\\
\bottomrule
\end{tabular}
\caption{Statistical overview of the CoRe dataset showing number of hypotheses, articles, triplets along with the distribution of labels across various topics within the dataset. Columns \emph{Train, Dev., Test} correspond to the number of triplets within each respective split.\\\emph{Hyp.=Hypotheses; Art.=Articles; Tri.=Triplets; Ent.=Entail; Cont.=Contradict; Inc.=Inconclusive; Dev.=Development}}
\label{tab:dataset}
}

\end{table*}
\section{Dataset}
Our Collaborative Reviews (CoRe) dataset is built from 12 different open-source collaborative literature reviews actively curated and maintained by domain experts and focused on specific questions in the social and behavioral sciences~\citep{haidt2019there, haidt2023social, haidt2022social, haidt2022alt, haidt2022free, haidt2022gamble, haidt2022videogame, haidt2022genz, haidt2022ocah, haidt2022ccaact, haidt2022puah, haidt2022amds, haidt2022phonefree}. The majority of these reviews were started in 2019 to map important studies within social and behavioral sciences and were maintained using Google docs. These documents are openly available for public viewing and academic researchers in relevant domains can request edit access to make changes. Each review categorizes articles based on a set of research questions related to the topic and the outcomes of each study. Any discrepancies in the classification are resolved by the lead authors alongside a domain expert\footnote{Further detail about these reviews can be found at \url{https://jonathanhaidt.com/reviews/}}.  

Figure \ref{fig:docstructure} gives a schematic illustration of a block of reviews about social media and mental health~\citep{haidt2023social}. Most articles are peer-reviewed scientific papers, but several reviews also contain blog posts, news articles, books, and other reports. 
Our CoRe dataset includes only scientific publications. 

Raw data were compiled using all reviews available on July 1, 2023. Research questions, labels, and Digital Object Identifiers (DOIs) were extracted from the reviews through automatic parsing of the document text. DOIs not readily available within the reviews were manually extracted from the publication links provided in the review. We then queried Semantic Scholar (S2) using DOIs to collect article titles and abstracts. In cases where S2 did not have coverage for certain articles, we queried CrossRef (CR). For articles outside the coverage of both S2 and CR, titles and abstracts were collected manually from the publication webpage. Research questions were converted to declarative statements in order to match the structure of hypotheses in the NLI task. Study outputs were manually mapped into one of three classes: \textit{entailment}; \textit{contradiction}; or, \textit{inconclusive}. The curated dataset contains \textit{(hypothesis, abstract, label)} triplets where the \textit{abstract} contains the evidence required to test the \textit{hypothesis} and predict the \textit{label}. 

\begin{table}[h]
 \fontsize{8.4}{11}\selectfont{
 \begin{tabular}{|c|c|c|c|c|c|}
  \toprule
  \textbf{Dataset}&\textbf{Hyp.}&\textbf{Pre.}&\textbf{Size}&\textbf{Domain}\\
  \midrule
  \textit{CoRe}&10&194&637&Social Sciences\\
  \textit{SciFact}&12&194&1,409&Medicine/Biology\\
  \textit{SNLI}&7&12&570,152&Non Scientific\\
  \bottomrule
 \end{tabular}}
  \caption{Comparison of average number of words in hypotheses, premises, instance counts, and domains.\emph{Hyp.=Hypotheses; Pre.=Premises}}
  \label{tab:data_compare}
\end{table}

Table~\ref{tab:dataset} provides a list of topics covered by the 12 collaborative reviews and an overview of key statistics. The dataset contains 69 distinct hypotheses tested across 602 scientific articles and findings aligned to our 3 labels. In total, the dataset contains 638 triplets because a fraction of articles address multiple hypotheses. The \textit{entailment} class has greatest representation within the dataset; 
61.6\% of triplets represent articles that contain evidence in support of the corresponding hypothesis. The \textit{contradiction} class makes up 25.7\% of the triplets. The remaining 12.7\% are in the \textit{inconclusive} class with a mixed evidence. The distribution of articles across topics is imbalanced, with some reviews containing substantially more literature than others. 

Table~\ref{tab:data_compare} presents a comparison between the CoRe dataset and the SNLI, and SciFact datasets. Notably, as the CoRe, SciFact datasets use abstracts of scientific publications as premises, their lengths are longer compared to that in SNLI dataset. The mean hypotheses, premise lengths in CoRe dataset are similar to SciFact dataset however, their domains are different. For training and evaluating the models, we shuffled the dataset and split it into to training (70\% ), development (15\%), and held-out test (15\%) datasets.
\section{Methods}
{We evaluate two families of methods on the SHE task using the CoRe dataset: transfer learning models and zero- and few-shot LLMs. In the case of transfer learning models, we evaluate sentence pair classifiers based on pre-trained embeddings and Natural Language Inference models.
}

\subsection{Sentence pair classification based on pre-trained embeddings}
\begin{figure}[h]
 \centering
   \centering
   \includegraphics[height=7cm, keepaspectratio]{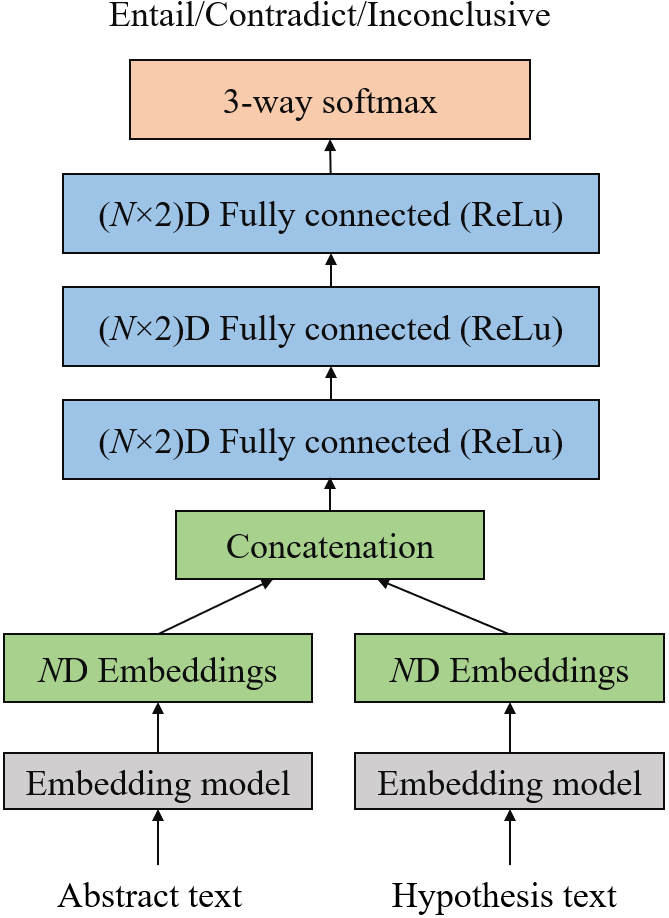}
   \caption{{Sentence pair classification based on pre-trained embeddings for concatenated hypothesis-abstract pairs}}
   \label{fig:mlparch}
\end{figure}

To investigate embedding models' performance on SHE, we adopt the sentence pair classification framework outlined in~\citep{bowman2015large}. Concatenated hypothesis and abstract embeddings are used as input to the model, which contains three successive fully-connected layers followed by a three-way softmax layer (see Figure \ref{fig:mlparch}). 

We evaluate the performance of two pre-trained embedding models: \textit{longformer}~\citep{beltagy2020longformer} and \textit{text-embedding-ada-002} \citep{openai2022ada}. Longformer is a transformer-based text encoder model developed to process information at the document-level, therefore eliminating the need for chunking long input text sequences. It uses a combination of local windowed attention and global attention to create a sparse attention matrix (vs. a full attention matrix) making attention more efficient. Longformer supports sequences of length up to 4,096 and produces embeddings of size 768.

Text-embedding-ada-002 is a transformer decoder language model developed by OpenAI and at the time of its release (December 2022) was shown to achieve state-of-the-art performance on tasks such as text search and sentence similarity~\citep{openai2022ada}. It is capable of embedding sequences of length up to 8,192 and generates 1,536-dimensional embedding vectors.
\subsection{Transfer learning using Natural Language Inference models}
In this approach, we treat SHE task as an NLI task. Specifically, we use an abstract as the premise and determine whether it entails a given hypothesis. Among models proposed for the NLI task, we evaluate the Enhanced Sequential Inference Model (ESIM)~\citep{Chen_2017} and Multi-Task Deep Neural Network (MT-DNN)~\citep{liu2019multitask}.

ESIM is a supervised learning model that uses bidirectional Long Short Term Memory (biLSTM) layers to encode hypothesis and premise for inference~\citep{Chen_2017}. It has achieved high performance on NLI tasks and a reported accuracy of 88.6\% on the SNLI dataset. The model uses a 840B token version of GloVe embeddings~\citep{pennington2014glove} for word representations. 

MT-DNN is a model aiming at learning robust representations across NLU tasks, such as text summarization, NLI and question answering~\citep{liu2019multitask}. MT-DNN achieved the new state-of-the-art performance on the SNLI and the SciTail datasets. We use the MT-DNN model built on the pre-trained \textit{bert-base-uncased} model~\citep{devlin2019bert} and fine-tuned it over 5 epochs for the task.

\subsection{Large language models}
We tested two LLMs, ChatGPT and PaLM 2~\citep{anil2023palm}, on our test split. For ChatGPT model, we used the API version of \textit{gpt-3.5-turbo} which offers a faster and significantly less expensive model than OpenAI's other GPT-3.5, GPT-4 models. From PaLM 2, we used the generative model \textit{text-bison-001} \cite{palm2url} an LLM fine-tuned to follow natural language instructions on a variety of language tasks, e.g., information extraction, problem solving, text edition, and data extraction \cite{palmapi}. We explored these models' performance in zero-shot and few-shot settings. 
Models were prompted with the abstract and the hypothesis embedded into predefined templates. Prompts contained specific instructions to generate a single output label. 
\begin{table}
 \fontsize{8.4}{11}\selectfont{
 \begin{tabular}{|c|c|c|c|c|c|}
  \toprule
  \textbf{CoRe}&\textbf{SNLI}&\textbf{P1}&\textbf{P2, P3, P5}&\textbf{P4}\\
  \midrule
  \textit{Entail}&Entail&true&Yes&e\\
  \textit{Contradict}&Contradict&false&No&c\\
  \textit{Inconclusive}&Neutral&neutral&Maybe&n\\
  \bottomrule
 \end{tabular}}
  \caption{Label map across datasets and prompts.}
  \label{tab:mapping}
\end{table}
\paragraph{Prompt engineering\\}
Prompt engineering refers to the task of finding the best prompt for an LLM in support of given task \citep{liu2023pre}. We experiment with five prompts used in prior work. All are \textit{prefix} prompts, i.e., prompt text comes entirely before model-generated text. Prompt templates and their sources are summarized in Table \ref{tab:prompts}. Depending on the prompt template, we requested LLMs return one of three sets of labels: \emph{(true, false, neutral)}; \emph{(yes, no, maybe)}; \emph{(entail, contradict, neutral)}. Table \ref{tab:mapping} maps each label to our canonicalized label set. Because prompts were queried without providing any training data, we refer this method zero-shot learning.
\begin{table*}[h]
\small{
  \centering
  \begin{tabular}{|l|l|p{0.72\textwidth}|}
    \toprule
     \textbf{Id} & \textbf{Source} & \textbf{Template} \\
    \midrule
    P1 & \citep{basmov2023chatgpt} & You are given a pair of texts. Say about this pair: given Text 1, is Text 2 true, false or neutral (you can’t tell if it’s true or false)? Reply in one word.
    Text 1: {\color{blue}{Abstract}}
    Text 2: {\color{blue}{Hypothesis}}\\
    \midrule
    P2 & \citep{luo2023chatgpt}* & Decide if the following summary is consistent with the corresponding article. Note that consistency means all information in the summary is supported by the article.
Article: {\color{blue}{Abstract}}
Summary: {\color{blue}{Hypothesis}}
Answer (yes or no or maybe):\\
\midrule
  P3 & \citep{cheng2023uprise} & Here is a premise: ``{\color{blue}{Abstract}}." Here is a hypothesis: ``{\color{blue}{Hypothesis}}." Is it possible to conclude that if the premise is true, then so is the hypothesis? Yes, No, or Maybe?\\
  \midrule
  P4 & \citep{liu2023evaluating} & Instructions: You will be presented with a premise and a hypothesis about that premise. You need to decide whether the hypothesis is entailed by the premise by choosing one of the following answers: ’e’: The hypothesis follows logically from the information contained in the premise. ’c’: The hypothesis is logically false from the information contained in the premise. ’n’: It is not possible to determine whether the hypothesis is true or false without further information. Read the passage of information thoroughly and select the correct answer from the three answer labels. Read the premise thoroughly to ensure you know what the premise entails. Premise: {\color{blue}{Abstract}} Hypothesis: {\color{blue}{Hypothesis}}\\
  \midrule
  P5 & \citep{sun2023pushing}* & The task is to identify whether the premise entails the hypothesis. Please respond with ”yes” or ”no” or "maybe"
Premise: {\color{blue}{Abstract}}
Hypothesis: {\color{blue}{Hypothesis}}
Answer:\\
\bottomrule
\multicolumn{3}{l}{* Added a third label \textit{maybe}}
  \end{tabular}
  \caption{Overview of the different prompts used for testing LLMs and their sources. Since prompts P2, P5 have only two labels \textit{yes, no}, a third label \textit{maybe} was added.}
  \label{tab:prompts}}
\end{table*}

\paragraph{Prompt ensembling\\}
In the context of LLMs, prompt ensembling refers to using several individual prompts at inference~\citep{liu2023pre}. Ensembling has shown better performance than fine-tuned models by harnessing their complementary strengths~\citep{li2023making}. Here, we use a majority voting strategy to ensemble the outputs of our five individual prompts.

\paragraph{Few Shot Learning\\}
In the few-shot learning (FSL) setting, LLMs are provided with several examples that demonstrate how the model should respond to the prompt~\citep{brown2020language}. Studies show that the examples chosen for FSL have a significant impact on the performance of LLMs~\citep{kang2023large}. We used a semantic search method to select nine samples from the training dataset to provide examples for each hypothesis-abstract pair in the held-out dataset.\footnote{Number of training samples was constrained by the LLM prompt length limitations.} 
\begin{figure}[h]
 \centering
 \includegraphics[width=\linewidth]{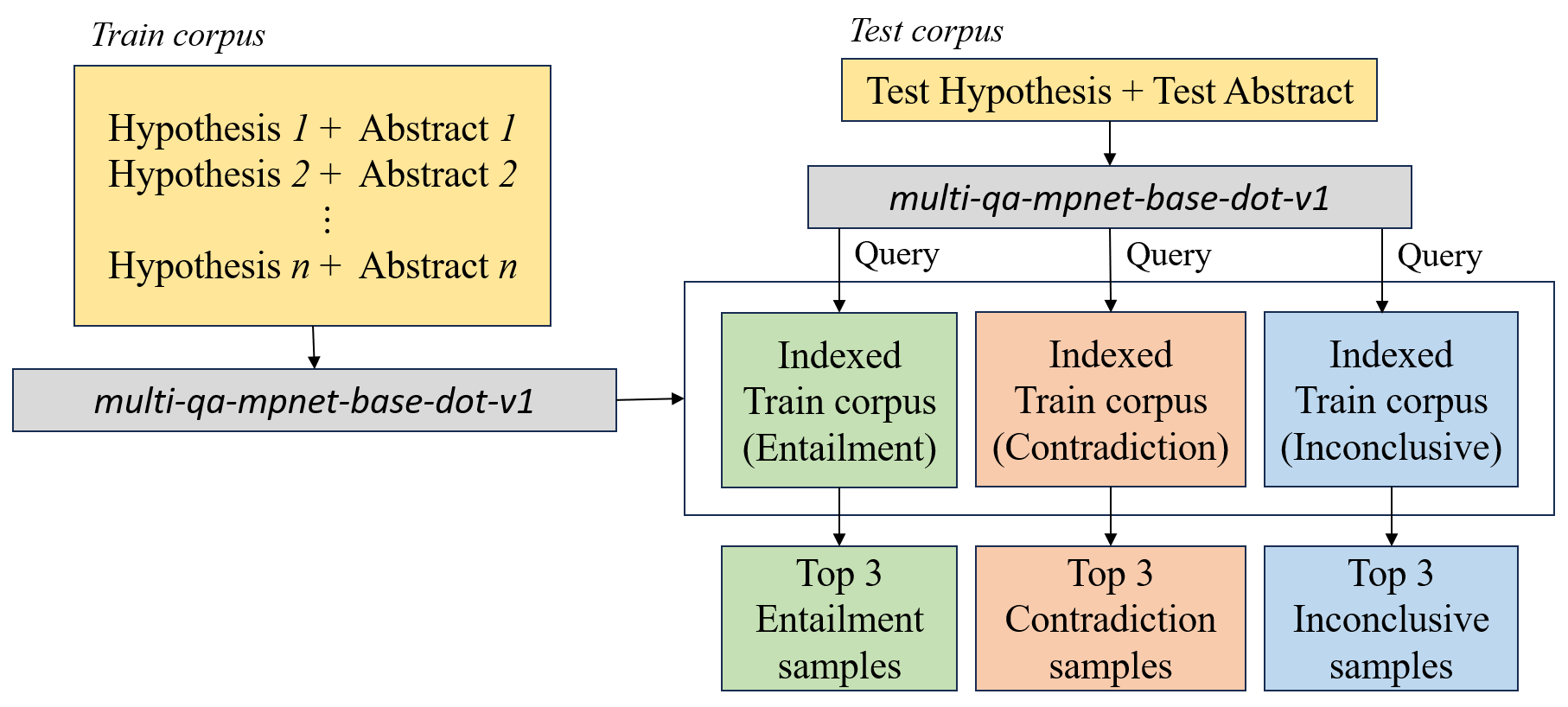}
 \caption{Semantic search-based sample selection for few-shot learning.}
 \label{fig:SampleSelection}
\end{figure}

To do so, we incorporated a pre-trained transformer encoder model, specifically the Huggingface implementation of \textit{multi-qa-mpnet-base-dot-v1}\footnote{\url{https://huggingface.co/sentence-transformers/multi-qa-mpnet-base-dot-v1}} which was designed for semantic search. As shown in Figure \ref{fig:SampleSelection}, we first split the training set into three subsets each having a different label. For each instance in the test dataset, we calculate cosine similarity between this instance against each concatenated hypothesis-abstract vectors in the training set and selected the top three pairs in each subset. This results in 9 hypothesis-abstract pairs used for FSL. For each instance in the held-out set, we calculated cosine similarity between the concatenated hypothesis-abstract pair and examples in the training corpora. 

\section{Experiments}
We evaluate each model's ability to discern the relationship between a given abstract and a hypothesis, written as a declarative statement in the CoRe dataset. Our approach aligns with methodologies widely used in the literature to evaluate performance on NLI datasets, e.g., SNLI \citep{bowman2015large}, MNLI \citep{williams2017broad}, where models are presented with a premise and a declarative hypothesis asked to classify their relationship. Performance of all models is measured by macro-F1-score, calculated as the average of F1-scores over all three class labels, and accuracy is calculated as the fraction of correct predictions. 

For sentence pair classification based on embeddings, all layers utilize the ReLU activation function. Hyperparameters such as learning rate and regularization parameter, were set based on Bayesian hyperparmeter tuning with an objective to maximize macro-F1-score on the test data~\citep{akiba2019optuna}. For training ESIM and fine-tuning MT-DNN on the SNLI dataset, we adopted default hyperparameters, as recommended in respective papers.

We evaluate LLMs in zero-shot and few-shot settings. The temperature parameter controls the creativity of the text generated by the LLMs. Lower temperatures result in more consistent output and higher temperatures result in more creative, diverse responses. We compare the performance of LLMs with temperature settings from 0 to 1, by increments of 0.25. We query each LLM using the same prompt 5 times in each temperature setting. 

To test prompt ensembling, we query each LLM using the set of five prompts and determine the final classification label by majority voting aggregation. Prompt ensembling was tested in both zero-shot and few-shot settings across different temperatures. Similarly to the single prompt evaluation, we tested the prompt ensembling under five independent runs for each temperature configuration. 

\section{Results}
Table \ref{tab:results} summarizes model performance on the test set. Here, we focus on comparing different types of models, so we report metrics averaged across all settings. The observation that all models achieve macro-F1-scores less than 0.65 demonstrates that SHE is a challenging task. \textbf{The sentence pair classifier model using \textit{text-embedding-ada-002} embeddings yielded the best performance achieving a macro-F1-score of 0.615}, followed by the pre-trained gpt-3.5-turbo model with prompt ensembling in the few-shot setting. 

\subsection{Natural Language Inference models}
As anticipated, ESIM and MT-DNN models when trained or fine-tuned on the SNLI dataset respectively, exhibited significantly lower performance compared to the model when trained or fine-tuned on the CoRe dataset. This can be attributed to the differences in the characteristics of hypotheses and premises in the two datasets. For instance in CoRe, each hypothesis has an average length of 10 words compared to 7 in case of SNLI. Average context length is 194 in CoRe compared to a 13-word premise in SNLI. Furthermore, levels of abstraction of hypotheses within the datasets vary. In CoRe, the evidence required to identify the abstract-hypothesis relationship is latent within the premise. Additionally, around 2,500 words from the CoRe dataset vocabulary are missing from the SNLI dataset. This underscores the need for domain specific datasets for fine-tuning.

\begin{table*}
\centering
\fontsize{9.5}{11}\selectfont{
\begin{tabular}{|p{3cm}|l|l|l|l|}
  \toprule
  \textbf{Type} & \textbf{Model} & \textbf{Setting} & \textbf{Accuracy} & \textbf{macro F1} \\
  \midrule
  \multirow{2}{3cm}{{Sentence pair classification}} & Longformer & Supervised on CoRe & 65.60\% & 0.558 \\
  \cmidrule{2-5}
  & text-embedding-ada-002 & Supervised on CoRe & \textbf{70.31}\% & \textbf{0.615} \\
  \midrule
  \multirow{4}{3cm}{{Transfer learning using NLI models}} & \multirow{2}{*}{MT-DNN} & Fine-tuned on CoRe & 67.97\% & 0.523 \\
   & & Fine-tuned on SNLI & 42.97\% & 0.342 \\
   \cmidrule{2-5}
  & \multirow{2}{*}{ESIM} & Supervised on CoRe & 64.84\% & 0.489 \\
  & & Supervised on SNLI & 39.84\% & 0.335 \\
  \midrule
  \multirow{8}{*}{LLM}& \multirow{4}{*}{ChatGPT} & Zero-shot w/o ensemble & 47.22\%* & 0.414* \\
   & & Few-shot w/o ensemble & 59.85\%* & 0.517*\\
   & & Zero-shot with ensemble & 53.94\% & 0.500\\
   & & Few-shot with ensemble & 66.57\% &0.576\\
   \cmidrule{2-5}
  & \multirow{4}{*}{PaLM 2} & zero-shot w/o ensemble & 59.78\%* & 0.504* \\
   & & Few-shot w/o ensemble & 69.78\%*\textsuperscript{\textdagger} & 0.583*\textsuperscript{\textdagger}\\
   & & Zero-shot with ensemble & 62.87\% & 0.536\\
   & & Few-shot with ensemble & 76.40\% & 0.678*\textsuperscript{\textdagger} \\ 
   \bottomrule
   \multicolumn{3}{l}{{*} Mean of responses across all temperatures, prompt templates, and iterations}\\
   \multicolumn{3}{l}{\textsuperscript{\textdagger} Incomplete responses}
\end{tabular}
\caption{Results summarizing the performance of models on the held-out set under different settings.}
\label{tab:results}}
\end{table*}
\subsection{Performance of LLMs}
\begin{figure}[h]
 \centering
 \begin{subfigure}[b]{1.1\linewidth}
   \includegraphics[width=\linewidth]{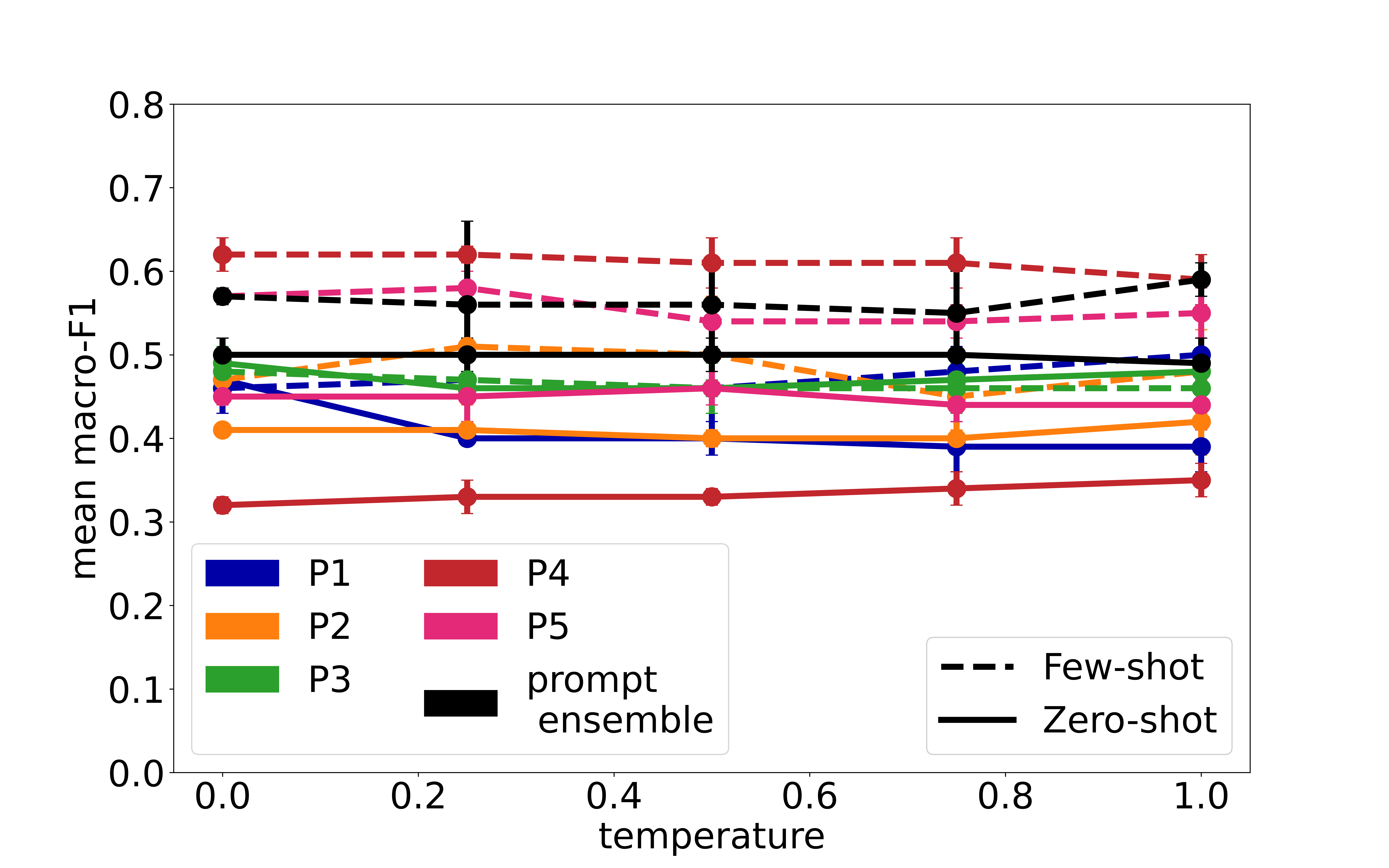}
   \caption{ChatGPT.}
   \label{fig:gpt_results}
  \end{subfigure}\\
  \begin{subfigure}[b]{1.1\linewidth}
   \includegraphics[width=\linewidth]{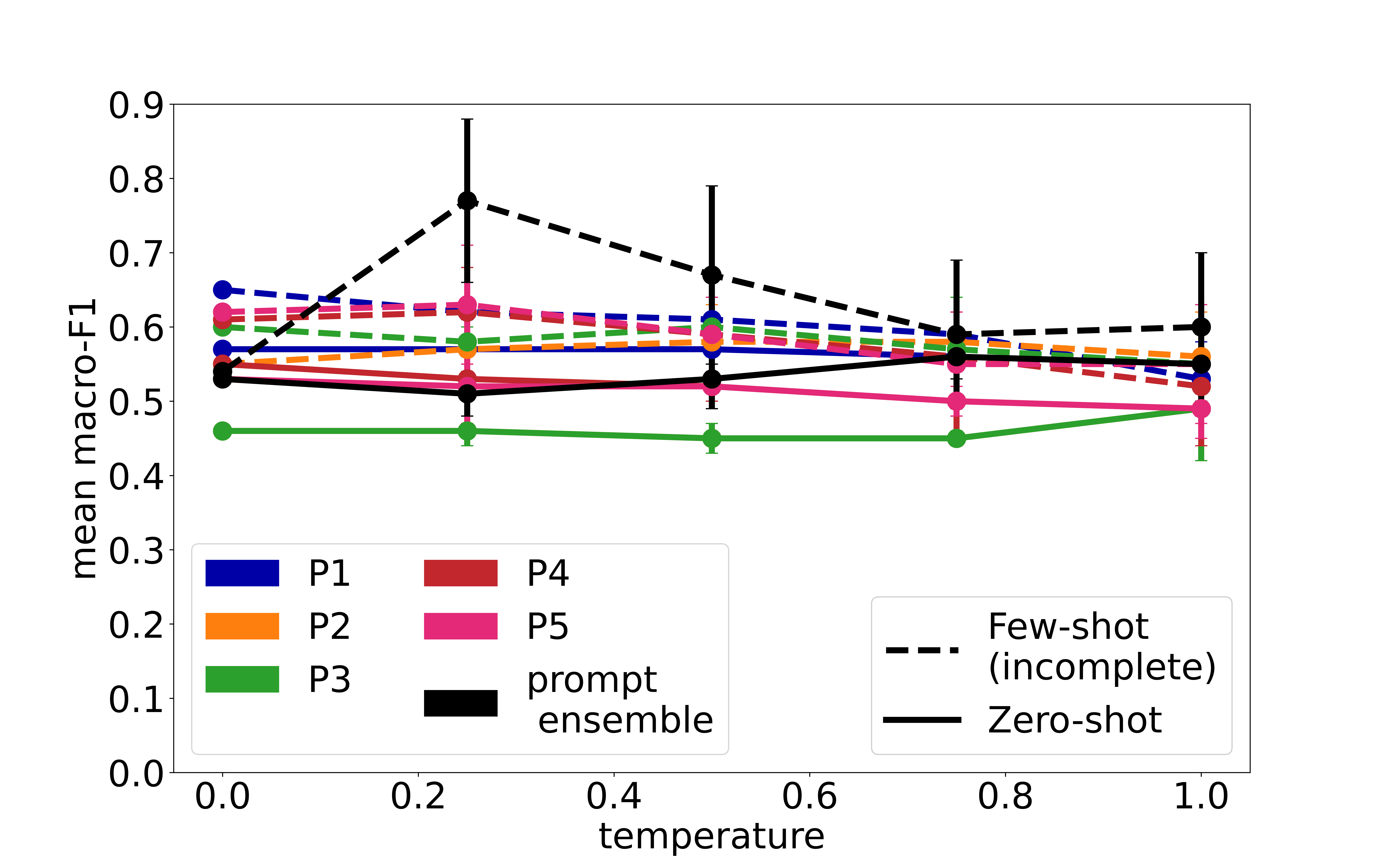}
   \caption{PaLM 2.}
   \label{fig:palm_results}
  \end{subfigure}
  \caption{Average macro-F1 of LLMs with different prompt templates and temperature settings.}
  \label{fig:summary_results}
\end{figure}

Figure \ref{fig:summary_results} summarizes the results for LLMs tested under different settings. Although not directly trained on the CoRe dataset, LLMs were able to comprehend the evidence within scientific abstracts and relate them to hypotheses. This can be attributed to their large-scale training. In the zero-shot setting, LLMs generally achieved macro-F1 of about0.5, which is comparable with transfer learning models fine-tuned on the CoRe dataset. Additionally, in the zero-shot setting across all prompt styles, PaLM 2 consistently outperformed ChatGPT.

We observed that when conducting FSL with PaLM 2, the model may output \textit{null} output and this occurrences are unpredictable at different temperatures and iterations. This occured over different temperatures and different iterations. The number of such instances ranged from 48 (37.5\%) to 64 (50\%) for an evaluation under a certain setting; overall, 20 (15.6\%) of responses yielded \textit{null} outputs consistently across all prompt styles. We will investigate this anomaly in the future.

While variations in performance were observed across individual runs, for a given prompt, the temperature setting did not have a major influence on the average performance of LLMs. This observation remained consistent across the zero-shot, few-shot, and prompt ensembling settings.
\paragraph{Effect of prompt template\\}
As expected, the choice of prompt style had clear influence on LLM performance. Different performance metrics for models tested with different prompts are summarized in Table~\ref{tab:prompt_results}. In the zero-shot setting, ChatGPT recorded the lowest mean macro-f1-score of 0.337 when prompted with P4 whereas it achieved highest mean macro-f1-score of 0.479 when using prompt P3. A similar trend is observed in the few-shot setting where the lowest performance was recorded when using prompt P3 with a mean macro-f1-score of 0.466 while the highest performance was reported for P4 with a mean macro-f1-score of 0.609. There was no single prompt that had consistent high performance across all temperatures, models, and settings. This indicates that ensembling over multiple prompt templates is more reliable than using a single prompt. 

\paragraph{Unequal performance gains with FSL \\}
FSL in general enhances the performance of LLMs across prompt styles, although performance gains are unequal. In case of ChatGPT, prompt P4 showed the greatest improvement with mean macro-f1-score increasing from 0.337 in zero-shot to 0.609 in the few-shot setting. Conversely, the performance slightly declined with P3, from 0.479 in zero-shot to 0.466 in the few-shot setting. Prompt ensembling with ChatGPT in the zero-shot setting achieves performance comparable to the few-shot setting.

\begin{table}
\centering
 \fontsize{8.3}{11}\selectfont{
 \begin{tabular}{|c|c|c|c|c|}
  \toprule
  \textbf{Model}&\textbf{Prompt}&\textbf{F1}&\textbf{P}&\textbf{R}\\
  \midrule
   \multirow{5}{*}{\shortstack{PaLM 2\\Zero-shot}}&\textit{P1}&0.56\textsubscript{(0.01)}&0.60\textsubscript{(0.01)}&0.59\textsubscript{(0.01)}\\
  &\textit{P2}&0.46\textsubscript{(0.02)}&0.50\textsubscript{(0.03)}&0.51\textsubscript{(0.01)}\\
  &\textit{P3}&0.46\textsubscript{(0.02)}&0.50\textsubscript{(0.03)}&0.51\textsubscript{(0.01)}\\
  &\textit{P4}&0.52\textsubscript{(0.02)}&0.60\textsubscript{(0.06)}&0.59\textsubscript{(0.04)}\\
  &\textit{P5}&0.51\textsubscript{(0.01)}&0.53\textsubscript{(0.01)}&0.53\textsubscript{(0.01)}\\
  \midrule
  \multirow{5}{*}{\shortstack{PaLM 2\\Few-shot*}}&\textit{P1}&0.60\textsubscript{(0.04)}&0.65\textsubscript{(0.03)}&0.68\textsubscript{(0.06)}\\
  &\textit{P2}&0.57\textsubscript{(0.01)}&0.58\textsubscript{(0.04)}&0.58\textsubscript{(0.01)}\\
  &\textit{P3}&0.58\textsubscript{(0.02)}&0.61\textsubscript{(0.02)}&0.64\textsubscript{(0.03)}\\
  &\textit{P4}&0.58\textsubscript{(0.04)}&0.58\textsubscript{(0.04)}&0.59\textsubscript{(0.04)}\\
  &\textit{P5}&0.59\textsubscript{(0.04)}&0.63\textsubscript{(0.07)}&0.59\textsubscript{(0.03)}\\
  \midrule
  \multirow{5}{*}{\shortstack{ChatGPT\\Zero-shot}}&\textit{P1}&0.41\textsubscript{(0.03)}&0.54\textsubscript{(0.032)}&0.50\textsubscript{(0.02)}\\
  &\textit{P2}&0.41\textsubscript{(0.01)}&0.44\textsubscript{(0.03)}&0.47\textsubscript{(0.00)}\\
  &\textit{P3}&0.47\textsubscript{(0.01)}&0.58\textsubscript{(0.02)}&0.53\textsubscript{(0.02)}\\
  &\textit{P4}&0.34\textsubscript{(0.01)}&0.61\textsubscript{(0.03)}&0.44\textsubscript{(0.01)}\\
  &\textit{P5}&0.45\textsubscript{(0.01)}&0.51\textsubscript{(0.01)}&0.50\textsubscript{(0.01)}\\
  \midrule
  \multirow{5}{*}{\shortstack{ChatGPT\\Few-shot}}&\textit{P1}&0.47\textsubscript{(0.02)}&0.47\textsubscript{(0.03)}&0.50\textsubscript{(0.01)}\\
  &\textit{P2}&0.49\textsubscript{(0.02)}&0.52\textsubscript{(0.02)}&0.53\textsubscript{(0.03)}\\
  &\textit{P3}&0.4\textsubscript{(0.01)}&0.56\textsubscript{(0.013)}&0.53\textsubscript{(0.01)}\\
  &\textit{P4}&0.61\textsubscript{(0.01)}&0.63\textsubscript{(0.01)}&0.64\textsubscript{(0.01)}\\
  &\textit{P5}&0.56\textsubscript{(0.03)}&0.57\textsubscript{(0.02)}&0.58\textsubscript{(0.02)}\\
  \bottomrule
  \multicolumn{5}{l}{* Partial results due to \textit{null} responses}
 \end{tabular}}
  \caption{Comparison of performance metrics of LLMs across various prompt templates on the CoRe dataset. The metrics are averaged for different temperature settings across all the runs. Subscripts indicate standard deviation over 5 runs. \emph{Note: The macro averaged precision and recall metrics are skewed due to class imbalance.}}
  \label{tab:prompt_results}
\end{table}
\section{Conclusion}
We have introduced the Scientific Hypothesis Evidencing task and a novel dataset for this task. Our goal is to determine whether a paper, based on its abstract, offers evidence in support or refute of a given hypothesis. This goal broadly underlies all of meta-analysis. It supports efforts to highlight inconsistencies and gaps in existing literature, motivate next studies, and support evidence-based decision-making and policy. Given the wide availability of abstracts, e.g., in scholarly data repositories, methods which can successfully operate on abstracts as opposed to full text are preferable.

Our CoRe dataset exhibits imbalance in class distribution and topical coverage. Imbalance in label classes is likely to be pervasive given publication bias where positive outcomes are more likely to be published than negative or inconclusive results. This imbalance is likely to affect the performance of models requiring training or fine-tuning, e.g., sentence pair classification, ESIM, and MT-DNN models. However, the impact on the performance of LLMs and pre-trained models, which are used solely for inference, is expected to be lesser.

Our findings suggest that hypothesis evidencing is challenging task for current NLU models, including state-of-the-art LLMs trained on a diverse set of data. Notably, sentence pair classification using embedding-based models outperforms LLMs in our experiments. Yet, these models are known to be less generalizable than LLMs.

Looking ahead, Open AI has recently introduced fine tuning functionality for gpt-3.5-turbo. Future work should investigate the performance of LLMs following fine tuning on the data. This will indicate whether the higher performance of embedding-based models is result of exposure to the complete training dataset vs. fewer examples provided in FSL setting.
In addition, for a more robust assessment, future work should explore five-fold cross-validation. And, given the limitations associated with human-generated discrete prompts, future work should explore automatic prompt tuning. 

We expect that, in providing the research community with an initial benchmark dataset, our work will catalyze some of these next steps. Future work should continue to build out datasets like ours, keeping in mind ways to ameliorate class imbalance and diversify represented topics. {Furthermore, future work should prioritize moving beyond binary or ternary labels towards appropriately capturing nuances in hypothesis-evidence relationships}. Notably, the assembly of these datasets serves multiple aims, bringing together domain experts around important questions in their own area, highlighting robustness and reliability amongst claims, and of course moving forward NLP and NLU.

\section*{Ethics Statement}
The data we compiled were originally contributed by volunteering social scientists. Although we did not seek their consensus, we cite each review as an individual reference and acknowledge all contributors for their efforts on this open list, which not only benefits social scientists but also computer and information scientists. In addition, because the edit request needed to be approved, we assumed the contributors were all qualified scientists or researchers, so the quality of the data was ensured.

\section{Bibliographical References} 
\bibliographystyle{lrec-coling2024-natbib}
\bibliography{lrec-coling2024-example}
\end{document}